\documentclass[letterpaper]{article} % DO NOT CHANGE THIS
\usepackage{fgnerf}  % DO NOT CHANGE THIS
\usepackage{times}  % DO NOT CHANGE THIS
\usepackage{helvet}  % DO NOT CHANGE THIS
\usepackage{courier}  % DO NOT CHANGE THIS
\usepackage[hyphens]{url}  % DO NOT CHANGE THIS
\usepackage{graphicx} % DO NOT CHANGE THIS
\usepackage{natbib}  % DO NOT CHANGE THIS AND DO NOT ADD ANY OPTIONS TO IT
\usepackage{caption} % DO NOT CHANGE THIS AND DO NOT ADD ANY OPTIONS TO IT
\usepackage{amsmath}
\usepackage{multirow}
\usepackage{booktabs}
\usepackage{amsfonts} 
\usepackage{soul}
\usepackage{xcolor}
\usepackage{algorithm}
\usepackage{algorithmic}
\usepackage{newfloat}
\usepackage{listings}

\DeclareCaptionStyle{ruled}{labelfont=normalfont,labelsep=colon,strut=off} % DO NOT CHANGE THIS
\floatstyle{ruled}
\newfloat{listing}{tb}{lst}{}
\floatname{listing}{Listing}
\nocopyright % -- Your paper will not be published if you use this command
\title{FG-NeRF: Flow-GAN based Probabilistic Neural Radiance Field for Independence-Assumption-Free Uncertainty Estimation}
\author{
    Songlin Wei\equalcontrib\textsuperscript{\rm 1,2}
    Jiazhao Zhang\equalcontrib\textsuperscript{\rm 1,2}
    Yang Wang\textsuperscript{\rm 1}
    Fanbo Xiang\textsuperscript{\rm 3}
    Hao Su\textsuperscript{\rm 3}
    He Wang\textsuperscript{\rm 1,2}
}
\affiliations{
    \textsuperscript{\rm 1}CFCS, School of Computer Science, Peking University\\
    \textsuperscript{\rm 2}Beijing Academy of Artificial Intelligence\\
    \textsuperscript{\rm 3}UC San Diego\\
}

\usepackage{bibentry}
\begin{document}

\maketitle
\begin{abstract}
Neural radiance fields with stochasticity have garnered significant interest by enabling the sampling of plausible radiance fields and quantifying uncertainty for downstream tasks. Existing works rely on the independence assumption of points in the radiance field or the pixels in input views to obtain tractable forms of the probability density function. However, this assumption inadvertently impacts performance when dealing with intricate geometry and texture.
In this work, we propose an independence-assumption-free probabilistic neural radiance field based on Flow-GAN. By combining the generative capability of adversarial learning and the powerful expressivity of normalizing flow, our method explicitly models the density-radiance distribution of the whole scene. 
We represent our probabilistic NeRF as a mean-shifted probabilistic residual neural model. Our model is trained without an explicit likelihood function, thereby avoiding the independence assumption.
Specifically, We downsample the training images with different strides and centers to form fixed-size patches which are used to train the generator with patch-based adversarial learning.
Through extensive experiments,  our method demonstrates state-of-the-art performance by predicting lower rendering errors and more reliable uncertainty on both synthetic and real-world datasets.
%\st{Moreover, we build an active NeRF reconstruction application based on our probabilistic NeRF learning framework to demonstrate the high quality of our uncertainty estimation.}
\end{abstract}
\section{Introduction}
As a popular approach for neural scene modeling, Neural Radiance Fields (NeRF)~\cite{nerf} have been widely studied in recent years due to their ability to render photorealistic novel views. Additionally, NeRF has shown effectiveness in various related fields, including 3D reconstruction~\cite{wang2021neus, yariv2021volume, instant-ngp}, camera-pose recovery~\cite{gnerf, schwarz2020graf}, 3D semantic segmentation~\cite{zhi2021place, Vora2021NeSFNS}, and photo-realistic simulation~\cite{Li2023PACNeRFPA, Li20213DNS}. However, the majority of existing works~\cite{Gao2022NeRFNR} only output one color for each pixel while falling short to capture the uncertainty in neural scene modeling. 
This limitation hinders the applications of NeRFs in active perception and interaction tasks, \textit{e.g.}, robotics~\cite{ran2022neurar, active-nerf}, autonomous driving~\cite{Li2022READLN}, and human-computer interaction~\cite{Li2022RTNeRFRO}, where it is essential to incorporate uncertain information~\cite{Cai2021CuriositybasedRN}.

\begin{figure}[t]
    \centering
    \includegraphics[width=1.0\linewidth]{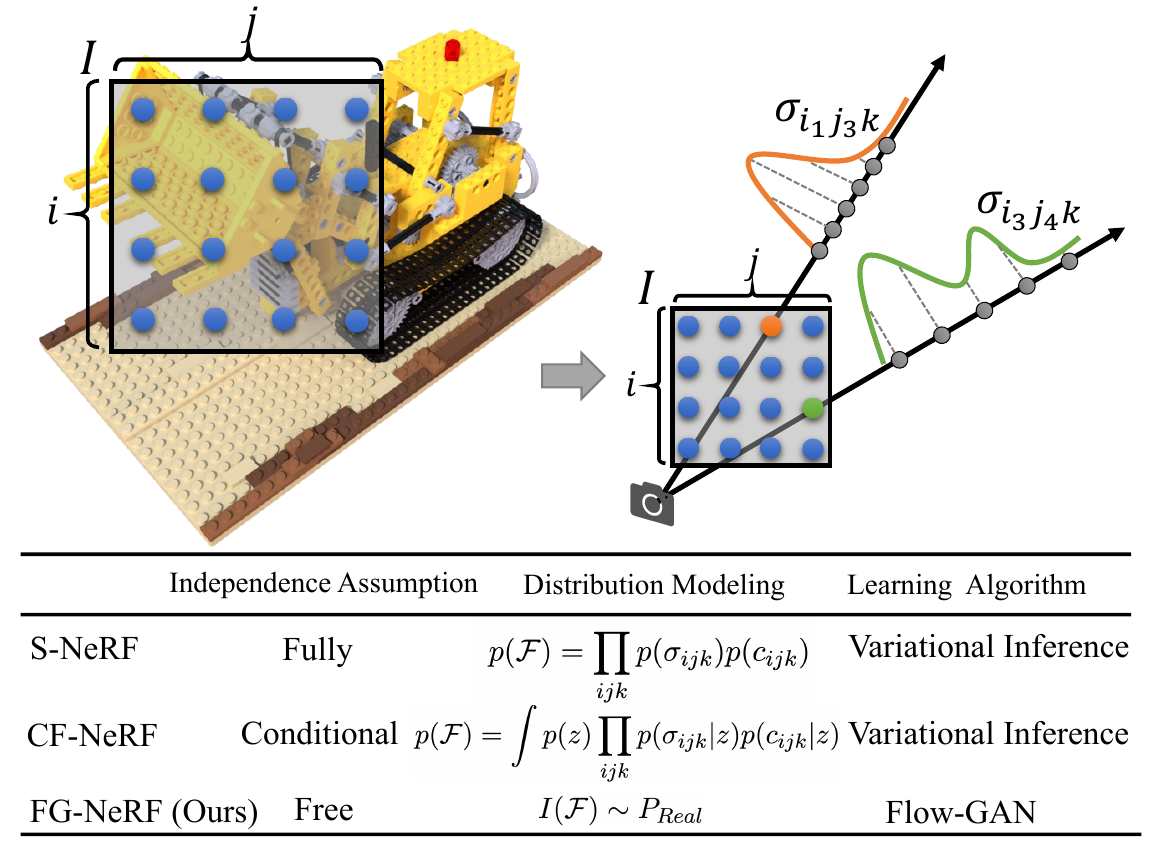}
    \caption{We propose to use adversarial learning to model probabilistic neural radiance field $\mathcal{F}$. Previous work models the whole density-radiance probability distribution as $p(\mathcal{F})$. S-NeRF factorizes the $p(\mathcal{F})$ as the product of the density distribution $p(\sigma_{ijk})$ and radiance distribution $p(c_{ijk})$ of all sampled points, while CF-NeRF \cite{cf-nerf} uses latent variable $z$ to make the sampled points conditional independent. On the contrary, our work FG-NeRF directly models the distribution $P_{Real}$ of the rendered image $I$ through volume rendering of $\mathcal{F}$.}
    \label{fig_teaser}
\end{figure}

To capture the uncertainty of NeRF, recent studies have proposed various types of probabilistic neural radiance fields~\cite{s-nerf, cf-nerf} that aim to learn a distribution over the radiance field.
However, learning a distribution over such continuous fields presents significant challenges, due to the vast probability space involved.
Consequently, existing approaches usually rely on strong assumptions.
For example, Stochastic NeRF (S-NeRF)~\cite{s-nerf} 
assumes the radiance and density values are independent Gaussian distributions at different locations, which neglects the important spatial consistency in 3D spaces.
The state-of-the-art method, Conditional-Flow NeRF (CF-NeRF)~\cite{cf-nerf}, leverages normalizing flow~\cite{cnf} technique to capture arbitrary radiance/density distributions. Furthermore, CF-NeRF introduces a global latent variable and achieves conditional independence among the values via De Finetti's representation theorem \cite{kirsch2019elementary}.
However, we contend that the conditional independence assumption still restricts the capacity of probabilistic NeRF to represent complex and large scenes accurately.

In this work, we propose a novel probabilistic neural radiance field, Flow-GAN NeRF (FG-NeRF). Inspired by Flow-GAN~\cite{flow-gan}, our method leverages both normalizing flow~\cite{cnf} and GAN~\cite{goodfellow2020generative}.
We propose to use adversarial training to learn  the radiance/density distributions of the whole scene.
In our FG-NeRF, the normalizing flow module serves only as a generator and is trained along with an additional discriminator, which together form a GAN framework.
% \st{Unlike S-NeRF and CF-NeRF, which maximizes the probability of the color for each pixel independently, our GAN loss supervises the colors from all interested rays together and thus is free from any independence assumption.
% Moreover, our normalizing flow is not conditional on a global latent variable and hence is more computationally efficient and expressive.}
As a result, our method exhibits two primary distinctions from S-NeRF or CF-Nerf (See Fig. \ref{fig_teaser}: \textit{first}, we leverage generative adversarial learning instead of variational inference used in prior methods. \textit{Second}, our method dispenses with the need for scene-wide distribution factorization to compute the likelihood or self-entropy term during variational inference (See Sec. \ref{eqn3}), thereby achieving independence-assumption free.
Consequently, FG-NeRF demonstrates the capability to produce high-quality uncertainty estimations while effectively capturing intricate appearance and geometry in room-level scenes.

Through extensive experiments, FG-NeRF showcases new state-of-the-art performance when compared to existing uncertainty estimation methods on LLFF~\cite{nerf}, ScanNet~\cite{scannet}, and Replica~\cite{replica} datasets. Our proposed method achieves improved uncertainty estimation, accompanied by high-quality color and depth predictions. Moreover, by integrating FG-NeRF with a multi-level hash encoding technique~\cite{instant-ngp}, our approach achieves a remarkable four times faster compared to CF-NeRF. With this significant improvement in computational efficiency, we further build an autonomous NeRF reconstruction, showcasing the superior performance of our predicted uncertainty.

\section{Related Work}
NeRF~\cite{nerf} has been widely studied due to its impressive capability in reconstructing scenes for photorealistic novel views. By utilizing a sparse collection of 2D views and employing a neural network to encode a representation of the 3D scene. This neural network consists of a set of fully-connected layers that output the radiance for any given input 3D spatial location and 2D viewing direction within the scene. In recent years, NeRF has been extensively researched to improve the quality~\cite{zhang2020nerf++, kosiorek2021nerf, barron2021mip, wang2021neus} and efficiency~\cite{neff2021donerf, instant-ngp, chen2022tensorf}. However, despite the advancements made, NeRF is unable to quantify the uncertainty associated with rendered views or estimated geometry, which limits its applicability in active tasks. To address this limitation, we propose FG-NeRF, a probabilistic neural radiance field that explicitly quantifies uncertainty.

\paragraph{Uncertainty estimation in NeRF.}
The estimation of uncertainty has been a widely explored topic in deep learning~\cite{hendrycks2019using, brachmann2016uncertainty, malinin2018predictive, teye2018bayesian, geifman2018bias}. Here, we only review the papers that are highly relevant to NeRF. Early studies~\cite{lee2022uncertainty} regard uncertainty as the entropy of the density among the sampled ray. Despite the simplicity, this approach tends to perform poorly on complex geometry and lacks the capability to estimate uncertainty at arbitrary positions. 
Another line of research~\cite{ran2022neurar, active-nerf, nerfw} attempted to implicitly predict the variance associated with rendered pixels or positions. However, These methods apply the volume rendering process over uncertainty values are not theoretically grounded.

Recently, the most compelling approach is to learn a posterior distribution over the model parameters. S-NeRF~\cite{s-nerf}, for instance, models a distribution encompassing all possible radiance fields that explain the scene. To ensure tractability, S-NeRF makes a strong assumption by assuming that all predicted distributions of each point are independent. In an effort to alleviate this limitation, CF-NeRF~\cite{cf-nerf} softens this assumption by conditioning the predicted distributions on a shared latent variable~\cite{yang2019pointflow}. This design enables spatially smooth distribution and improves the quality of predictions. However, we argue that CF-NeRF still relies on a conditional independence assumption through the use of a global latent variable.
Notably, our FG-NeRF stands out by sidestepping the independence assumptions by employing a generator and a discriminator that are trained adversarially.

\paragraph{Adversarial Learning for NeRF.}
Adversarial learning~\cite{goodfellow2020generative, creswell2018generative, grover2018flow} has recently emerged as a promising approach for learning the radiance field for various challenging tasks. Previous relevant studies~\cite{gnerf, schwarz2020graf, chan2021pi} have primarily focused on leveraging generative models to learn the radiance field from unstructured 2D images.
Recent research efforts~\cite{deng2022gram, chan2022efficient} have made significant advancements in enhancing synthesis quality by achieving higher resolutions, improved 3D geometry, and faster rendering. However, to the best of our knowledge, our work is the first to utilize adversarial learning specifically for uncertainty estimation in NeRF. 
\section{Background}
\label{background}
\paragraph{Deterministic and probabilistic formulations of NeRF}
Neural radiance field (NeRF) \cite{nerf} is an implicit scene representation. It is modeled as a continuous function $\mathcal{F}:(x,d) \rightarrow (\sigma, c)$, which maps the pair of 3D spatial location $x \in \mathbb{R}^3$ and viewing direction $d \in \mathbb{S}^2 $ into density $\sigma \in \mathbb{R}^+$ and RGB radiance $c \in [0,1]^3 $.
Given a camera ray $r(t)=x_0+td ~$shooting from origin $x_o$ along the direction $d$, the color $c$ of the ray is computed via the physics-based volume rendering:
\begin{equation} \label{eqn1}
\begin{aligned}
  & c(r) = \int_{0}^{\infty} T(t) \sigma(r(t)) c(r(t), d)dt, \\
  & ~\textnormal{where}~ T(t)=exp(-\int_{0}^{t} \sigma(r(t))ds),
\end{aligned}
\end{equation}
and the function $T(t)$ is the accumulated transmittance along the ray. It is the probability that the ray travels from the origin to $x_t$ without hitting any obstacles. 

The radiance field $\mathcal{F}$ is \textit{deterministic} in terms of having a single density and RGB radiance for each point in the space. Deterministic NeRF is unable to output the uncertainty of the rendering when the output confidence is desired. One approach to remedy this problem is to formulate the neural radiance construction as \textit{probabilistic}. Probabilistic NeRF models the scene as a joint probabilistic density and radiance distribution $p(\mathcal{F})$ for all the spatial locations and view directions. 
In this manner, the deterministic radiance field can be regarded as one realization over the distribution $p(\mathcal{F})$.

\paragraph{Modeling density-radiance distributions with normalizing flow}
\label{subsec_flow}
To learn the probability distribution $\mathcal{F}$ of the whole scene, we first discuss how to model the distribution of a single point through normalizing flow.
For any location $x$ viewed from direction $d$, the conditional joint distribution of the density and radiance is $p(\sigma,c|x,d)=p(\sigma|x)p(c|x,d)$ where density $\sigma$ is assumed to depend only on location $x$.
Approximating the radiance distribution $p(c|x,d)$ with unimodal distributions like Gaussian is sub-optimal, as a point in the real-world can have different colors viewed from different directions due to reflection or iridescence.
Inspired by CF-NeRF \cite{cf-nerf}, we use conditional normalizing flow (CNF) \cite{cnf} which transforms a simple distribution into arbitrary complex shape conditioning on the location-view pair $(x,d)$. 
Specifically, conditional normalizing flow maps the initial random variable $(\sigma_0,c_0)$ with the prior distributions $q_\sigma(\sigma_0)$ and $q_c(c_0)$ into the target $(\sigma,c)$ via $K$ steps of bijective mappings $\sigma_k = f_k^\sigma(\sigma_{k-1},x)$ and $c_{k} = f_k^c(c_{k-1},x,d)$ respectively. The target distributions can be computed via the change-of-variables formula:
\begin{equation} \label{eqn6}
  p(\sigma|x) = q_\sigma(\sigma_0) \prod_{k=1}^{K-1} \left\vert \det{ \frac{\partial f_k^\sigma }{\partial\sigma_{k-1}}} \right\vert ^{-1},
\end{equation}
\begin{equation} \label{eqn7}
  p(c|x,d) = q_c(c_0) \prod_{k=1}^{K-1} \left\vert \det{ \frac{\partial f_k^c }{\partial c_{k-1}}} \right\vert ^{-1}
\end{equation}
The initial distribution $q_\sigma$ and $q_c$ is usually set to standard Gaussian distributions.

Because directly computing the density-radiance distribution $p(\mathcal{F})$ for the whole scene is intractable, previous work \cite{s-nerf}\cite{cf-nerf} approximate the probabilistic distribution using variational inference \cite{vi} via a parametric function $q(\mathcal{F};\theta)$.
The models boil down to the computation of a likelihood term $p(\mathcal{T}|\mathcal{F})$ of the training sample and a self-entropy term.
They assume the distribution for each point in the scene is independent or conditionally independent to constrain the $q(\mathcal{F};\theta)$, which can then be factorized into the product of each point distribution to make the computation tractable:

\begin{equation} \label{eqn3}
  q_{\vartheta}(\mathcal{F}) = \prod_{x \in \mathbb{R}^3} \prod_{d \in \mathbb{S}^2} p(\sigma|x) p(c|x,d),~~ \textnormal{or}
\end{equation}
\begin{equation} \label{eqn3}
  q_{\vartheta}(\mathcal{F}|z) = \int{ p(z) \prod_{x \in \mathbb{R}^3} \prod_{d \in \mathbb{S}^2} p(\sigma|x,z) p(c|x,d,z)}.
\end{equation}
% However, neighboring points in the scene are likely to have similar density or color, which even makes the conditional independence assumption questionable.
%
Nevertheless, neighboring 3D points within the scene are prone to exhibit analogous density or color, thereby casting doubt on the plausibility of the conditional independence assumption.

Although these works produce high-quality results of uncertainty estimation on object-level scenes, experiments show that this assumption hampers the performance when the scene becomes large and complex.

\section{Method}
\label{method}
\begin{figure*}
  \centering
  \includegraphics[width=0.958\linewidth]{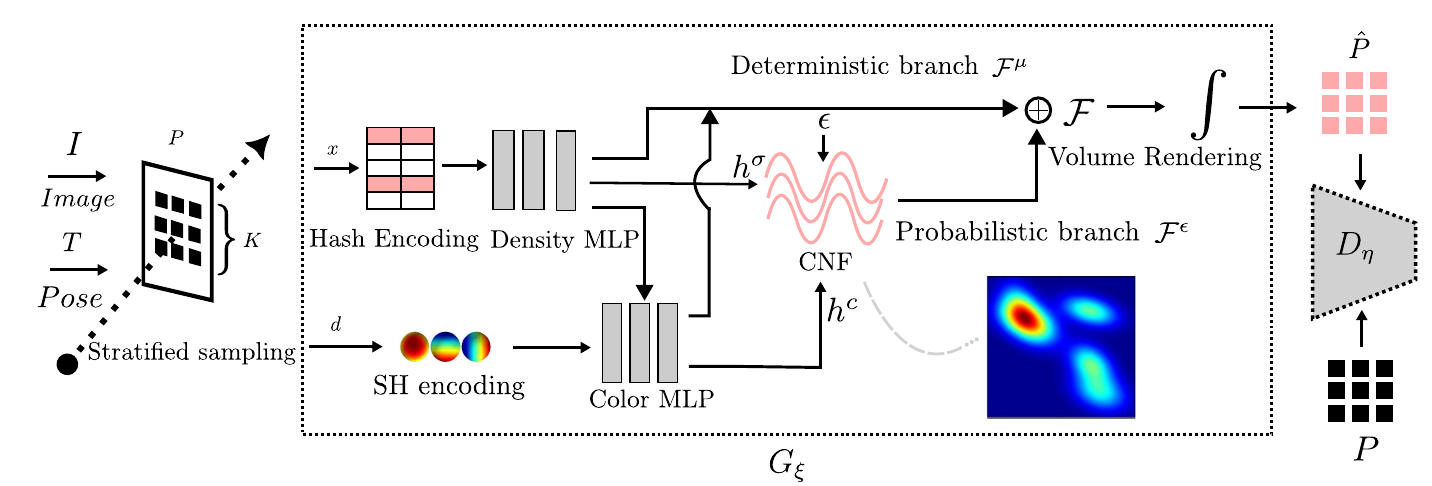}
  \caption{Overview of our pipeline. Our pipeline learns to generate fake image patches given the training image $I$ and pose $T$. The generator $G_\xi$ and discriminator $D_\eta$ jointly guide the normalizing flow (CNF) for modeling the density/radiance distribution.}
  \label{fig_pipeline}
\end{figure*}
We study the probability distribution of the entire density-radiance field. More concretely, given $\mathcal{T} = \{(I^i, T^i, D^i)\}_{i=1}$ denotes the triplets of image $I$, pose $T$ and optional depth $D$ indexed by $i$, we goal is to learn the posterior distribution $p(\mathcal{F}|\mathcal{T})$. 

Our whole pipeline is illustrated in Fig. \ref{fig_pipeline}. Given an image $I$, we dynamically sample a batch of rays $\mathcal{R}$ to obtain an image patch $P$. The location-view pairs $(x,d)$ which are sampled through the stratified sampling of the rays originating from camera pose $T$ are fed into the generator $G_\theta$. The generator consists of two branches: (1) the deterministic branch learns the mean values for (2) the probabilistic branch which, is implemented with the conditional normalizing flow (CNF), learns to predict the mean-shifted probability distribution of the radiance field of the scene. Finally, a synthesized patch $\hat{P}$ is obtained through volume rendering of the combined branches and is sent to the discriminator $D_\phi$ to compare with the real training patch $P$.

% The subsequent subsections are organized as follows: In Section \ref{subsec_decompose}, we present the decomposition method for the probabilistic neural radiance field. This is followed by Section \ref{subsec_flow_gan}, where we provide an extensive discussion on the adversarial learning process for the generator and discriminator. Together, they guide the normalizing flow to learn a more meaningful distribution. Section \ref{subsec_depth} delves into the utilization of depth supervision as a regularization technique for the normalizing flow. Once the density and radiance distributions have been learned, Section \ref{subsec_quantify} explains how uncertainty quantification is achieved. Lastly, in Section \ref{subsec_training}, we outline the training protocols employed in our approach.

\subsection{Decomposing probabilistic neural radiance field}
\label{subsec_decompose}
Directly learning the probability distribution of the density-radiance field presents additional challenges due to the significant divergences in distribution between occupied and free regions.
Consequently, the presence of large density values (ranging from $0$ to $+\infty$) can lead to numerical instability during the training of the normalizing flow.
To tackle this challenge, we decompose $\mathcal{F}$ into a deterministic branch and a residual probabilistic branch:
\begin{equation}
  \mathcal{F} = \mathcal{F}^\mu + \mathcal{F}^{\epsilon}
\end{equation}
The deterministic branch $\mathcal{F}^\mu $ acting like the mean of the probabilistic density-radiance field is estimated following the standard volume rendering procedure.
It provides the mean values of density and radiance for each location-view pair (x,d) in the scene.
Thus, the probabilistic branch $\mathcal{F}^\epsilon$ is only needed to learn a mean-shifted residual distribution. The distributions are learned via conditional normalizing flow as described in Section \ref{background}.
Given a batch of rays $\mathcal{R}$ sampled from the training images $I$,
the deterministic branch $\mathcal{F}^\mu$ of the network is supervised with the $L2$ loss between the rendered color $\hat{c}(r)$ and the ground truth color of each ray $r$. The loss is denoted as  
\begin{equation} \label{eqn2}
  \mathcal{L}_{det} = \sum_{r \in \mathcal{R}} \Vert I[r] - \hat{c}(r) \Vert ^2_2 ~
\end{equation}
where $\hat{c}(r)$ represents the rendered color of the ray $r$. It is obtained through alpha-composition:
\begin{equation} \label{eqn4}
\begin{aligned}
  & \hat{c} = \sum_{i=1}^{N} T^i \alpha^i c^{\mu,i}, \\
  & T^i = \prod_{j=1}^{i-1} (1-\alpha^j), \\
  & \alpha^i = 1 - \exp(-\sigma^{\mu,i} \delta^i)
\end{aligned}
\end{equation}
where superscript $i$ indicates the index of the discrete sampling points along the ray. $\alpha$ denotes the opacity of the color at the sampling point and $\delta$ is the distance between neighboring sampling points.
With this deterministic branch been regularized by the rendering loss, our network can efficiently learn a rough geometry and appearance of the scene. Furthermore, the residual mean-shifted probabilistic branch $\mathcal{F}^{\epsilon}$ is eased to learn a zero-mean centered distribution of density $\sigma^\epsilon$ and radiance $c^\epsilon$.

Thus, the combined density and radiance are
\begin{equation} \label{eqn8}
  \sigma = \sigma^\mu + \sigma^\epsilon ~~~~~~ c = c^\mu + c^\epsilon ~.
\end{equation}

\subsection{Adversarial learning of probabilistic NeRF based on Flow-GAN}
\label{subsec_flow_gan}
\begin{figure}[h]
  \includegraphics[width=0.9\linewidth]{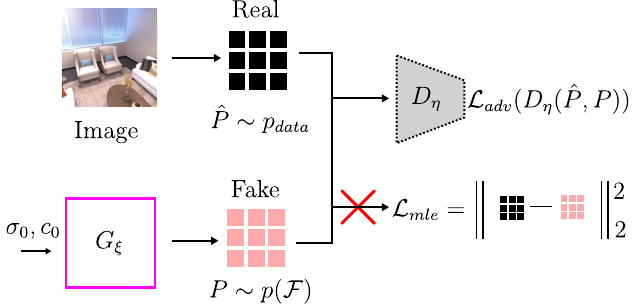} 
  \caption{Comparison of patch-based adversarial loss and pixel-level rendering loss}
  \label{fig_adv_mle}
\end{figure}

Instead of the factorizations of the probability distributions as done in \cite{s-nerf}\cite{cf-nerf}, we propose to employ adversarial learning to sidestep the direct computation of the likelihood functions as illustrated in Fig. \ref{fig_adv_mle}. 
Therefore, our method does not rely on any form of independence assumptions.
We first dynamically sample pixels \cite{patch-gan} structured with $K \times K$ pixels from training image $I$. The image patch follows the real data distribution $p_{data}$. Then, we train a generator $G_\xi$ to synthesize the patch using known camera pose $T$ and a discriminator $D_\eta$ to compare the real and fake patches.
More specifically, the $K^2$ rays in the patch are spaced evenly with random intervals and the center is also randomly shifted within the image. 
This sampling strategy preserves the image structure which allows the generator to generate high-quality samples.
We then employ stratified sampling following standard NeRF to get $N$ points along each ray. 
We follow \cite{instant-ngp} to use multi-resolution hash encoding to encode sampling points $x$ and use the spherical harmonics to encode the view direction $d$.

The generator $G_\xi$ is implemented with CNF. 
% Armed with the expressivity of the normalizing flow, we 
For each sampling point $x$, we first acquire its mean density $\sigma^\mu$ and color $c^\mu$ through the deterministic branch $\mathcal{F}^\mu$ and then transform an initial density $\sigma^\epsilon$ and color $c^\epsilon$ following standard Guassians into the target distribution using CNF, which is conditional on the location feature $h^\sigma$ and radiance feature $h^c$.
Subsequently, we synthesize the image patch via volume rendering of the combined densities and colors as in Eq. (\ref{eqn8}) and Eq. (\ref{eqn4}).
The synthesized image $\hat{P}$ will be fed into a discriminator $D_\eta$ to be compared with the real image patches $P$. 
Finally, we minimize a distribution distance between the generated image $p(\mathcal{F})$ and the real image patch $p_{data}$ based on adversarial loss $\mathcal{L}_{adv}$ as is done in classic GAN framework. The training objective is:
\begin{equation} \label{eqn_gan}
\begin{aligned}
  & \min_{\xi \in \Xi}  \max_{\eta \in E} \mathcal{L}_{adv} = \mathbb{E}_{P \sim p_{data}}[\log(D(P;\eta))] + \\
  & ~~~\mathbb{E}_{\hat{P} \sim p(\mathcal{F})}[\log(1-D(G(\sigma_0,c_0;\xi); \eta))].
\end{aligned}
\end{equation}
The discriminator $D_\eta$ is trained to output $1$ for real patches and $0$ for fake patches.
It is implemented as a convolutional neural network following \cite{schwarz2020graf}. The discriminator is conditioned on the scale $s$ to perform well. The scale $s$ is the ratio of the size of the sampled patch with respect to the input image.
$s$ is set to 1 at the beginning of the training to have a large perceptive field and gradually decreased to a small value to capture fine-grained details of the scene. 
The generator $G_\xi$ and discriminator $D_\eta$ are trained adversarially to promote the generator to synthesize more realistic images.
As a result, the normalizing flow in the generator is trained to produce high-quality density-radiance distributions for synthesizing photo-realistic images. 
Thus, uncertainty quantified by such distributions has better qualities compared with previous works as demonstrated in the experiments.

\subsection{Optional extra depth supervision}
\label{subsec_depth}
Unlike object-level scenes where training rays are densely available for every point in the scene. It is challenging for NeRF to model large scenes like ScanNet \cite{scannet} due to the limited number of training images.
In such cases, depth supervision further improves the learning of geometry and radiance. 
The ray depth can be obtained by the integral:
\begin{equation}
  \hat{D} = \sum_{i=1}^{N} w^i R^i
\end{equation}
where $w^i=T^i\alpha^i$ is the weight of the ray depth $R^i$ of the sampling point.
However, straightforward $L2$ loss $\mathcal{L}_{depth}=\vert\vert D-\hat{D}\vert\vert^2_2$ supervision performs poorly. 
We empirically found that using the cross-entropy loss between the sample weights $w(t) = T(t) \alpha(t)$ and an assumed ground truth depth centered Gaussian distribution $\mathcal{N}(D, \sigma^2)$ performs better than straightforward $L2$ loss. The cross-entropy loss is defined as:
\begin{equation}
  \mathcal{L}_{depth} = \frac{1}{\vert \mathcal{R} \vert} \sum_{r \in \mathcal{R}} \log w_r^i \exp(-\frac{(t_i-D)^2}{2\sigma^2})
\end{equation}
where $\Sigma$ is set to $0.1$ in the experiments. Intuitively, such cross entropy constrains the weights to follow a unimodal distribution hence enforcing stronger regularization when learning the density distributions. 

\subsection{Quantifying uncertainty}
\label{subsec_quantify}
After the distributions over the radiance field are learned, we are able to quantify the uncertainty of density or radiance of the scene. 
For each point $x$, we first sample $S$ values from the distribution learned with normalizing flow and then use the sample variance to quantify the density uncertainty. 
The density uncertainty is a measure of convergence of the learned density.
A similar procedure is employed to obtain the radiance uncertainty.
We can also quantify the uncertainty associated with the rendering image and depth map.
Given any camera pose, we run the generator $M$ times to sample $M$ trajectories of random variables $(x^i, c^i)_{1:M}$ along each ray.
Subsequently, we synthesize the images  $\hat{I}_{1:M}$ and depth maps $D_{1:M}$. 
The uncertainties can then be obtained through the color and depth variance of each pixel.
Object edges often have high-density uncertainties due to the discontinuity of the depth whereas areas with high-frequency colors often have high radiance uncertainties.
This kind of uncertainty is shown to be beneficial for active NeRF reconstruction in the experiments.

\subsection{Training protocol}
\label{subsec_training}

In contrast to standard NeRF, which primarily focuses on minimizing rendering loss, our GAN based on probabilistic NeRF incorporates a generator and discriminator that are trained in a min-max game framework. The generator comprises a deterministic branch and a probabilistic branch implemented with a normalizing flow. To ensure training stability, we adopt instance normalization \cite{ulyanov2016instance} and spectral normalization \cite{miyato2018spectral} in the discriminator. To enhance convergence, we employ a small number of $M=8$ samples for each point in the normalizing flow, gradually increasing it to 16 to generate $M$ fake image patches. Simultaneously, we replicate the real patch $M$ times before feeding it into the discriminator to maintain gradient balance during backpropagation. More implementation details can be found in the supplemental material.

\section{Experiments}
\label{experiments}
\paragraph{Datasets.}
We conducted evaluations of our method on three diverse publicly available datasets, namely LLFF \cite{nerf}, Replica \cite{replica}, and ScanNet \cite{scannet}. The LLFF dataset contains complex geometry and appearance, with camera views predominantly directed toward a single object. Notably, the LLFF dataset has also been utilized in S-NeRF~\cite{s-nerf} and CF-NeR~\cite{cf-nerf}. Additionally, we employed the Scannet and Replica datasets to evaluate the performance of our methods in challenging scenarios involving large scenes with diverse camera views. The Replica dataset comprises high-precision reconstructed scenes with nearly perfect rendering results, while the Scannet dataset consists of real-captured RGB-D sequences obtained using consumer-level cameras. Following the same experimental setting of \cite{Yu2022MonoSDFEM}, we utilized a subset of scenes in accordance with the methodology.

\begin{table*}[t]
    \caption{Comparison of uncertainty estimation and rendered color on the LLFF dataset. The best results are indicated in bold, with the second-best results underlined. }
    \label{tab_llff_rgb_uncertainty}
    \centering
    \resizebox{0.8\textwidth}{!}{
        \begin{tabular}{llccccccc}
            \toprule
            &  & \multicolumn{3}{c}{Quality Metrics} &  & \multicolumn{2}{c}{Uncertainty Metrics} &  \\ \cmidrule(r){3-5} \cmidrule(r){7-8}
            &  & PSNR$\uparrow$ & SSIM$\uparrow$ & LPIPS$\downarrow$ &  & AUSE RMSE$\downarrow$ & AUSE MAE$\downarrow$ & \\ 
            \midrule
            D.E. \cite{deep_ensemble} &  & \underline{22.32} & 0.788 & 0.236 &  & 0.0254 & 0.0122 &   \\
            Drop. \cite{mc_dropout} &  & 21.90 & 0.758 & 0.248 &  & 0.0316 & 0.0162 &   \\
            NeRF-W \cite{nerfw} &  & 20.19 & 0.706 & 0.291 &  & 0.0268 & 0.0113 &  \\
            S-NeRF \cite{s-nerf} &  & 20.27 & 0.738 & {0.229} &  &  0.0248 & {0.0101} &  \\
            CF-NeRF \cite{cf-nerf} &  & 21.96 & \underline{0.790} & \underline{0.201} &  & \underline {0.0177} & \underline{0.0078} &  \\ 
            FG-NeRF (Ours) & & \textbf{24.41} & \textbf{0.816} & \textbf{0.170} &  & \textbf{0.0118} & \textbf{0.0028} & \\
            \bottomrule
            \vspace{-10mm}
        \end{tabular}
    }
\end{table*}

\paragraph{Metrics.}
To evaluate the predicted uncertainty and rendered quality, we employ standard metrics commonly used in the literature~\cite{nerf, s-nerf, cf-nerf}. For uncertainty evaluation, we follow previous works by utilizing the AUSE (Area Under the Sparsification Error curve) metric proposed in~\cite{bae2021estimating, poggi2020uncertainty}. AUSE evaluates the correlation between uncertainty and predicted error. Given an error metric (RMSE or MAE), we can sort the pixels of the uncertainty map and the error map respectively. Repeatedly removing the top $1\%$ of the sorted error pixels, leads to a sparsification curve and an oracle curve. The AUSE is then computed as the area between these two curves.
For rendered quality evaluation, we employ widely adopted metrics such as PSNR, SSIM (Structural Similarity Index), and LIPIP (Learned Perceptual Image Patch Similarity). Detailed descriptions of these metrics can be found in the supplementary material.

% \paragraph{Implementation details.}
% The image is resized to $H \times W $=504$\times$378 on the LLFF \cite{nerf} dataset and 512 $\times$ 512 on ScanNet \cite{scannet} or Replica \cite{replica}. 
% We use a patch of $K \times K$=1024 rays and batch size $B$=1 to train the network.
% The $K^2$ rays are sampled evenly in the training image with scale $s$ annealing from 0.8 to $K/min(H,W)$.
% The number of sampling points along each ray is $N$=256 and the number of importance sampling is $N_f$=128.
% Besides, we sample $M$=32 points from the initial standard Gaussian distributions in the normalizing flow.
% We use RMSProp optimizer and set the learning rates of the generator and discriminator as $0.01$ and $0.005$ respectively.

\begin{figure}[t]
    \centering
    \includegraphics[width=1\linewidth]{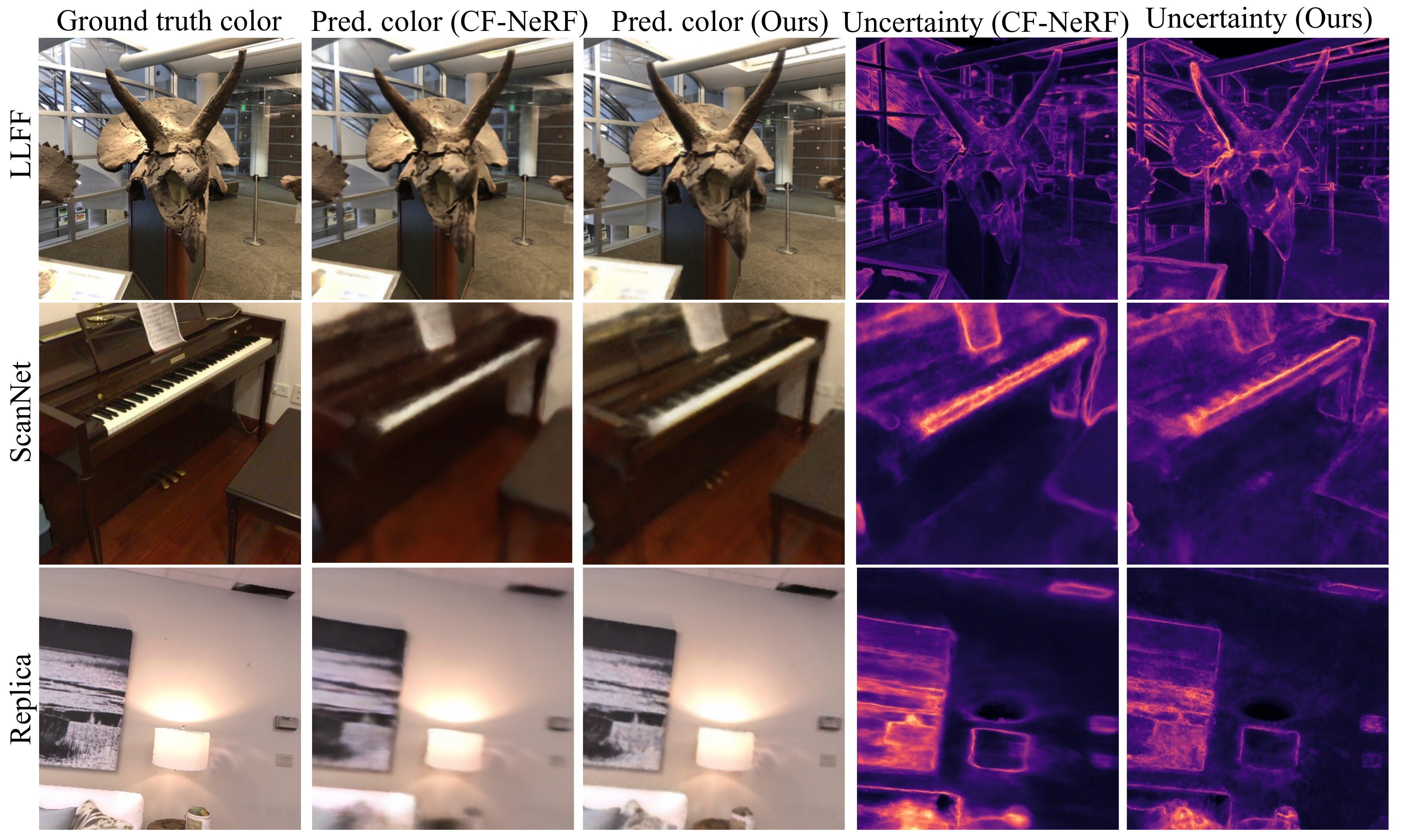}
    \caption{Quality comparison on the LLFF, ScanNet, and Replica datasets.}
    \label{fig_uncertainty_quality}
    \vspace{-5mm}
\end{figure}

\subsection{Comparision on LLFF dataset}
\label{exp_quality}
Following the same experimental setup in~\cite{cf-nerf}, we evaluate our uncertainty estimation method on the LLFF dataset to compare with existing methods \cite{deep_ensemble, mc_dropout, nerfw, cf-nerf,s-nerf}.
Note that, the camera views from LLFF dataset are primarily forward-facing to a central object within a small range of angles. Consequently, the coordinates can be constrained within the range of $[0,1]$ using Normalized Device Coordinates (NDC), thereby facilitating the learning of the radiance field.
The average results of eight scenes from LLFF dataset are presented in Tab. \ref{tab_llff_rgb_uncertainty}. 
Our method FG-NeRF achieves significant improvements in uncertainty and demonstrates better rendering quality as well (see Fig.~\ref{fig_uncertainty_quality} third rows for a visual comparison). Compared to the previous SOTA method CF-NeRF, the AUSE of our method decreased by $33.3\%$ and $64.1\%$ on RMSE and MAE, respectively, clearly demonstrating the superiority of our methods. Additionally, the results prove that the CF-NeRF outperforms the S-NeRF by relaxing the strong independence assumption. Furthermore, our method achieves the best performance by introducing adversarial learning~\cite{flow-gan} to sidestep any independence assumptions.

\begin{table*}[t]
    \caption{Quantitatively comparison with baseline method on Scanet and Replica datasets.}
    \label{tab_compare_cfnerf}
    \centering
    \resizebox{0.8\textwidth}{!}{
    \begin{tabular}{cllcccccc}
    \toprule
    \multirow{2}{*}{Scene Type}  &  \multirow{2}{*}{Method} & & \multicolumn{3}{c}{Quality Metrics} &  & \multicolumn{2}{c}{Uncertainty Metrics}   \\ 
    \cmidrule(r){4-6} \cmidrule(r){8-9}
    & & & PSNR$\uparrow$ & SSIM$\uparrow$ & LPIPS$\downarrow$ &  & AUSE RMSE$\downarrow$ & AUSE MAE$\downarrow$  \\ \midrule
    \multirow{3}{*}{ScanNet} & CF-NeRF \cite{cf-nerf} &  & 20.67 & 0.627 & 0.477 &  & 0.0293 & 0.0353   \\ 
    &NeurAR \cite{ran2023neurar} &  & 23.78 & 0.765 & \textbf{0.347} &  & 0.0265 & 0.0150  \\
    &Ours &  & \textbf{24.76} & \textbf{0.793} & 0.388 &  & \textbf{0.0136} & \textbf{0.0043}  \\
    \midrule
    \multirow{3}{*}{Replica} 
    & CF-NeRF \cite{cf-nerf} &  & 20.35 & 0.621 & 0.418 &  & 0.0196 & 0.0058   \\ 
    & NeurAR \cite{ran2023neurar} & & 23.31 & 0.715 & 0.395 &     & 0.0175 & 0.0050    \\ 
    &Ours &  & \textbf{25.40} & \textbf{0.816} & \textbf{0.364} &  & \textbf{0.0136} & \textbf{0.0040}  \\
    \bottomrule
    \end{tabular}
    }
\end{table*}

\subsection{Comparison on Replica and Scannet dataset}
Compared to object-level scenes, room-scale scenes present additional challenges due to their diverse camera views and complex geometry and appearance. In this context, we compare our methods with the previous state-of-the-art approach CF-NeRF \cite{cf-nerf} and NeurAR~\cite{ran2022neurar}, specifically designed for scene-level uncertainty estimation. The results can be found in Table \ref{tab_compare_cfnerf}, and visual comparisons are illustrated in Figure \ref{fig_uncertainty_quality}. Notably, our method consistently achieves the best performance on both datasets. It is worth mentioning that CF-NeRF exhibits a performance drop when transitioning from the LLFF dataset (as shown in Table \ref{tab_llff_rgb_uncertainty}) to the ScanNet and Replica datasets (as shown in Table \ref{tab_compare_cfnerf}). In contrast, our method maintains a substantial level of performance, as observed in Table \ref{tab_llff_rgb_uncertainty} on the LLFF dataset. 
%To evaluate the performance of depth prediction in real-world environments, we conducted experiments on the ScanNet dataset.
%The results, presented in Table \ref{tab:table_exp_scannet50_metrics}, demonstrate that FG-NeRF achieves the most accurate depth prediction across all metrics. We believe this performance improvement comes from two key techniques: firstly, FG-NeRF avoids the reliance on independence assumptions and effectively encodes complex appearance and geometry; secondly, the mean-shifted probabilistic residual network maintains the influence of fine-grained radiance information. 

\begin{figure*}
    \centering
    \includegraphics[width=0.9\textwidth]{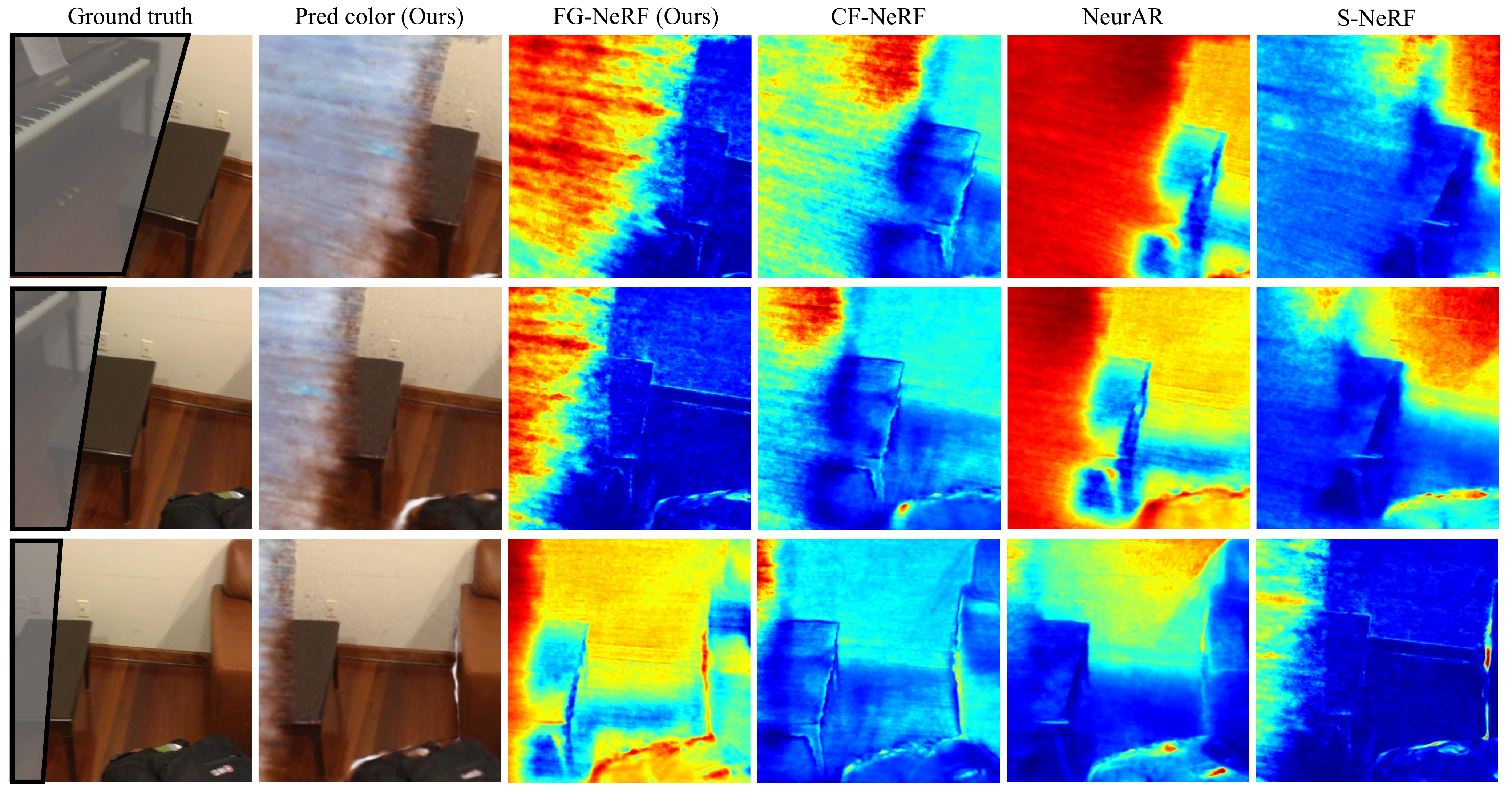}
    \caption{Comparison on uncertainty estimation under limited observations on ScanNet (the closer to blue, the smaller uncertainty). The unseen region is indicated by a gray mask.}
    \label{fig_uncertainty}
    \vspace{-2mm}
\end{figure*}

Additionally, we conduct a more challenging experiment to evaluate uncertainty estimation using a reduced number of frames ($10\%$ of the frames). 
% The presence of missing regions increases the difficulty in learning the radiance field from the observed regions while accurately predicting the boundary between the observed and unobserved regions (indicated by a gray mask). 
The visualizations can be found in Figure \ref{fig_uncertainty}, where it is evident that our method successfully predicts a clear boundary between the unobserved region and the observed space.
Moreover, our approach consistently maintains accurate uncertainty estimation along the object edges. 

\subsection{Ablation study}
%\label{ssec_hyper}
We conduct an ablation study to verify the effectiveness of key techniques and hyperparameter settings. 
As shown in Tab. \ref{tab_abl}, the baseline is our full implementation with adversarial learning and has a deterministic branch.
We first compare our proposed adversarial learning with plain L2 loss supervision as illustrated in Fig. \ref{fig_adv_mle}. The L2 loss is the squared difference of the pixel color between rendered image patch $\hat{P}$ and ground truth image patch $P$.
In the experiments, we observed that both the rendering quality and the uncertainty metrics of adversarial learning are better than the case with plain L2 loss supervision. 
We also ablation studies the influence of the deterministic branch as proposed in the main pipeline. Experiments have shown a slight performance drop when the deterministic branch is removed. 
Additionally, we studied the different settings of the hyperparameters, eg., the batch and image patch size, the sample size along each ray, and the scale annealing effect.
%The image is resized to $H \times W $=504$\times$378 on the LLFF \cite{nerf} dataset and 512 $\times$ 512 on ScanNet \cite{scannet} or Replica \cite{replica}. 
%We use a patch of $K \times K$=1024 rays and batch size $B$=1 to train the network.
%The $K^2$ rays are sampled evenly in the training image with scale $s$ annealing from 0.8 to $K/min(H,W)$.
%The number of sampling points along each ray is $N$=256 and the number of importance sampling is $N_f$=128.
%Besides, we sample $M$=32 points from the initial standard Gaussian distributions in the normalizing flow.

%We use RMSProp optimizer and set the learning rates of the generator and discriminator as $0.01$ and $0.005$ respectively.
%To enable a fair comparison with CF-NeRF \cite{cf-nerf}, we set the max epochs to match the total $100,000$ iterations in their code.
%An ablation study of our FG-NeRF with different settings of hyperparameters is presented in Tab. \ref{tab_abl}.

\begin{table}[t]
    \centering
    \caption{Ablations on LLFF Trex}
    \label{tab_abl}
    \resizebox{1\columnwidth}{!}{
    \begin{tabular}{lllcccccc}
        \toprule
        \multirow{2}{*}{} & \multirow{2}{*}{} & & \multicolumn{3}{c}{Quality Metrics} &  & \multicolumn{2}{c}{Uncertainty Metrics}   \\ 
        \cmidrule(r){4-6} \cmidrule(r){8-9} 
        & & & PSNR$\uparrow$ & SSIM$\uparrow$ & LPIPS$\downarrow$ &  & AUSE RMSE$\downarrow$ & AUSE MAE$\downarrow$  \\ 
        \midrule
        %\multirow{2}{*}{}  %$1 \times 32^2$  % Batch size, Patch size\\
        Baseline &  & & \textbf{26.58} & \textbf{0.891} & \textbf{0.105} &  & \textbf{0.0070} & \textbf{0.0017}   \\ 
        Color Supervision & L2 loss & & 25.89 & 0.842 & 0.197 &  & 0.0093 & 0.0043   \\ 
        Deterministic branch & off & & 24.96 & 0.723 & 0.285 &  & 0.0085 & 0.0049   \\ 
        %\midrule
         Batch $B$ $\times$ Patch $K^2  $  & $4 \times 16^2$ & & 25.43 & 0.817 & 0.091 &  & 0.0083 & 0.0024   \\
        %\midrule
        %\multirow{3}{*}{} 
        Sample size $M$& 24 & & 25.44 & 0.854 & 0.146 &  & 0.0079 & 0.0019   \\ 
        & 16 & & 24.64 & 0.825 & 0.184 &  & 0.0082 & 0.0019   \\
        & 8 & & 23.58 & 0.777 & 0.248 &  & 0.0089 & 0.0022   \\
        %& 32 & & 20.67 & 0.627 & 0.477 &  & 0.0293 & 0.0353   \\
        %\midrule
        %\multirow{2}{*}{Depth supervision} & MSE & & 22.89 & 0.742 & 0.297 &  & 0.0293 & 0.0353   \\ 
        %\midrule
        %\multirow{2}{*}{Scale annealing} & On & & 20.67 & 0.627 & 0.477 &  & 0.0293 & 0.0353   \\ 
        Scale annealing & Off & & 24.38 & 0.814 & 0.199 &  & 0.0087 & 0.0021   \\
        \bottomrule
    \vspace{-10mm}
    \end{tabular}
    }
\end{table}

\section{Limitations}
We acknowledge the presence of two main limitations in our approach. \textit{Firstly}, there is a significant computational cost requirement. FG-NeRG, similar to S-NeRF and CF-NeRF, is trained to predict a radiance/density distribution, which is more challenging to learn compared to a simplistic radiance/density value. Consequently, these methods require more extensive computational resources than the original NeRF framework. Despite adopting a multi-level hashing coding method~\cite{instant-ngp} for acceleration, our approach still takes nearly 6 GPU hours (NVIDIA V100) to obtain fully trained results on the LLFF dataset. This issue can potentially be alleviated by integrating advanced NeRF acceleration techniques~\cite{chen2022tensorf, Chen2022MobileNeRFET}. \textit{Secondly}, the rendering quality of our method is relatively low when compared to the latest works in NeRF literature. We are keen on further enhancing the quality by leveraging scene prior information~\cite{Peng2020ConvolutionalON}, incorporating advanced training strategies~\cite{Chen2022AugNeRFTS}, and employing novel backbone architectures~\cite{Sun2022Neural3R}.
\section{Conclusions}
In this paper, we propose FG-NeRF, a novel probabilistic neural radiance field, for high-quality uncertainty estimation. Our approach utilizes adversarial learning to model radiance/density distributions. Specifically, FG-NeRF performs as a generator and is trained in conjunction with a discriminator, forming an adversarial loss.
By jointly supervising the colors of all pixels within an image patch, our method avoids making the independence assumption. 
Consequently, FG-NeRF effectively learns intricate radiance and density distributions and demonstrates its capacity to predict high-quality uncertainty, radiance, and density.
Experimental evaluations on both synthetic and real-world datasets showcase that our method outperforms existing methods, exhibiting lower rendering errors and providing more reliable uncertainty predictions. Moreover, we build and evaluate an active NeRF reconstruction framework based on FG-NeRF, showcasing the potential applications of this method.
\bibliography{fgnerf}

\begin{thebibliography}{56}
\providecommand{\natexlab}[1]{#1}

\bibitem[{Bae, Budvytis, and Cipolla(2021)}]{bae2021estimating}
Bae, G.; Budvytis, I.; and Cipolla, R. 2021.
\newblock Estimating and exploiting the aleatoric uncertainty in surface normal
  estimation.
\newblock In \emph{Proceedings of the IEEE/CVF International Conference on
  Computer Vision}, 13137--13146.

\bibitem[{Barron et~al.(2021)Barron, Mildenhall, Tancik, Hedman,
  Martin-Brualla, and Srinivasan}]{barron2021mip}
Barron, J.~T.; Mildenhall, B.; Tancik, M.; Hedman, P.; Martin-Brualla, R.; and
  Srinivasan, P.~P. 2021.
\newblock Mip-nerf: A multiscale representation for anti-aliasing neural
  radiance fields.
\newblock In \emph{Proceedings of the IEEE/CVF International Conference on
  Computer Vision}, 5855--5864.

\bibitem[{Blei, Kucukelbir, and McAuliffe(2017)}]{vi}
Blei, D.~M.; Kucukelbir, A.; and McAuliffe, J.~D. 2017.
\newblock Variational inference: A review for statisticians.
\newblock \emph{Journal of the American statistical Association}, 112(518):
  859--877.

\bibitem[{Brachmann et~al.(2016)Brachmann, Michel, Krull, Yang, Gumhold
  et~al.}]{brachmann2016uncertainty}
Brachmann, E.; Michel, F.; Krull, A.; Yang, M.~Y.; Gumhold, S.; et~al. 2016.
\newblock Uncertainty-driven 6d pose estimation of objects and scenes from a
  single rgb image.
\newblock In \emph{Proceedings of the IEEE conference on computer vision and
  pattern recognition}, 3364--3372.

\bibitem[{Cai et~al.(2021)Cai, Chen, Wang, Zhang, and
  Meng}]{Cai2021CuriositybasedRN}
Cai, K.; Chen, W.; Wang, C.; Zhang, H.; and Meng, M.~Q. 2021.
\newblock Curiosity-based Robot Navigation under Uncertainty in Crowded
  Environments.
\newblock \emph{IEEE Robotics and Automation Letters}, 8: 800--807.

\bibitem[{Chan et~al.(2022)Chan, Lin, Chan, Nagano, Pan, De~Mello, Gallo,
  Guibas, Tremblay, Khamis et~al.}]{chan2022efficient}
Chan, E.~R.; Lin, C.~Z.; Chan, M.~A.; Nagano, K.; Pan, B.; De~Mello, S.; Gallo,
  O.; Guibas, L.~J.; Tremblay, J.; Khamis, S.; et~al. 2022.
\newblock Efficient geometry-aware 3D generative adversarial networks.
\newblock In \emph{Proceedings of the IEEE/CVF Conference on Computer Vision
  and Pattern Recognition}, 16123--16133.

\bibitem[{Chan et~al.(2021)Chan, Monteiro, Kellnhofer, Wu, and
  Wetzstein}]{chan2021pi}
Chan, E.~R.; Monteiro, M.; Kellnhofer, P.; Wu, J.; and Wetzstein, G. 2021.
\newblock pi-gan: Periodic implicit generative adversarial networks for
  3d-aware image synthesis.
\newblock In \emph{Proceedings of the IEEE/CVF conference on computer vision
  and pattern recognition}, 5799--5809.

\bibitem[{Chen et~al.(2022{\natexlab{a}})Chen, Xu, Geiger, Yu, and
  Su}]{chen2022tensorf}
Chen, A.; Xu, Z.; Geiger, A.; Yu, J.; and Su, H. 2022{\natexlab{a}}.
\newblock Tensorf: Tensorial radiance fields.
\newblock In \emph{Computer Vision--ECCV 2022: 17th European Conference, Tel
  Aviv, Israel, October 23--27, 2022, Proceedings, Part XXXII}, 333--350.
  Springer.

\bibitem[{Chen et~al.(2022{\natexlab{b}})Chen, Wang, Fan, and
  Wang}]{Chen2022AugNeRFTS}
Chen, T.; Wang, P.; Fan, Z.; and Wang, Z. 2022{\natexlab{b}}.
\newblock Aug-NeRF: Training Stronger Neural Radiance Fields with Triple-Level
  Physically-Grounded Augmentations.
\newblock \emph{2022 IEEE/CVF Conference on Computer Vision and Pattern
  Recognition (CVPR)}, 15170--15181.

\bibitem[{Chen et~al.(2022{\natexlab{c}})Chen, Funkhouser, Hedman, and
  Tagliasacchi}]{Chen2022MobileNeRFET}
Chen, Z.; Funkhouser, T.~A.; Hedman, P.; and Tagliasacchi, A.
  2022{\natexlab{c}}.
\newblock MobileNeRF: Exploiting the Polygon Rasterization Pipeline for
  Efficient Neural Field Rendering on Mobile Architectures.
\newblock \emph{ArXiv}, abs/2208.00277.

\bibitem[{Creswell et~al.(2018)Creswell, White, Dumoulin, Arulkumaran,
  Sengupta, and Bharath}]{creswell2018generative}
Creswell, A.; White, T.; Dumoulin, V.; Arulkumaran, K.; Sengupta, B.; and
  Bharath, A.~A. 2018.
\newblock Generative adversarial networks: An overview.
\newblock \emph{IEEE signal processing magazine}, 35(1): 53--65.

\bibitem[{Dai et~al.(2017)Dai, Chang, Savva, Halber, Funkhouser, and
  Nie{\ss}ner}]{scannet}
Dai, A.; Chang, A.~X.; Savva, M.; Halber, M.; Funkhouser, T.; and Nie{\ss}ner,
  M. 2017.
\newblock ScanNet: Richly-annotated 3D Reconstructions of Indoor Scenes.
\newblock In \emph{Proc. Computer Vision and Pattern Recognition (CVPR), IEEE}.

\bibitem[{Deng et~al.(2022)Deng, Yang, Xiang, and Tong}]{deng2022gram}
Deng, Y.; Yang, J.; Xiang, J.; and Tong, X. 2022.
\newblock Gram: Generative radiance manifolds for 3d-aware image generation.
\newblock In \emph{Proceedings of the IEEE/CVF Conference on Computer Vision
  and Pattern Recognition}, 10673--10683.

\bibitem[{Gal and Ghahramani(2016)}]{mc_dropout}
Gal, Y.; and Ghahramani, Z. 2016.
\newblock Dropout as a bayesian approximation: Representing model uncertainty
  in deep learning.
\newblock In \emph{international conference on machine learning}, 1050--1059.
  PMLR.

\bibitem[{Gao et~al.(2022)Gao, Gao, He, Lu, Xu, and Li}]{Gao2022NeRFNR}
Gao, K.; Gao, Y.; He, H.; Lu, D.; Xu, L.; and Li, J. 2022.
\newblock NeRF: Neural Radiance Field in 3D Vision, A Comprehensive Review.
\newblock \emph{ArXiv}, abs/2210.00379.

\bibitem[{Geifman, Uziel, and El-Yaniv(2018)}]{geifman2018bias}
Geifman, Y.; Uziel, G.; and El-Yaniv, R. 2018.
\newblock Bias-reduced uncertainty estimation for deep neural classifiers.
\newblock \emph{arXiv preprint arXiv:1805.08206}.

\bibitem[{Goodfellow et~al.(2020)Goodfellow, Pouget-Abadie, Mirza, Xu,
  Warde-Farley, Ozair, Courville, and Bengio}]{goodfellow2020generative}
Goodfellow, I.; Pouget-Abadie, J.; Mirza, M.; Xu, B.; Warde-Farley, D.; Ozair,
  S.; Courville, A.; and Bengio, Y. 2020.
\newblock Generative adversarial networks.
\newblock \emph{Communications of the ACM}, 63(11): 139--144.

\bibitem[{Grover, Dhar, and Ermon(2018{\natexlab{a}})}]{flow-gan}
Grover, A.; Dhar, M.; and Ermon, S. 2018{\natexlab{a}}.
\newblock Flow-gan: Combining maximum likelihood and adversarial learning in
  generative models.
\newblock In \emph{Proceedings of the AAAI conference on artificial
  intelligence}, volume~32.

\bibitem[{Grover, Dhar, and Ermon(2018{\natexlab{b}})}]{grover2018flow}
Grover, A.; Dhar, M.; and Ermon, S. 2018{\natexlab{b}}.
\newblock Flow-gan: Combining maximum likelihood and adversarial learning in
  generative models.
\newblock In \emph{Proceedings of the AAAI conference on artificial
  intelligence}, volume~32.

\bibitem[{Hendrycks et~al.(2019)Hendrycks, Mazeika, Kadavath, and
  Song}]{hendrycks2019using}
Hendrycks, D.; Mazeika, M.; Kadavath, S.; and Song, D. 2019.
\newblock Using self-supervised learning can improve model robustness and
  uncertainty.
\newblock \emph{Advances in neural information processing systems}, 32.

\bibitem[{Isola et~al.(2017)Isola, Zhu, Zhou, and Efros}]{patch-gan}
Isola, P.; Zhu, J.-Y.; Zhou, T.; and Efros, A.~A. 2017.
\newblock Image-to-image translation with conditional adversarial networks.
\newblock In \emph{Proceedings of the IEEE conference on computer vision and
  pattern recognition}, 1125--1134.

\bibitem[{Kirsch(2019)}]{kirsch2019elementary}
Kirsch, W. 2019.
\newblock An elementary proof of de Finetti’s theorem.
\newblock \emph{Statistics \& Probability Letters}, 151: 84--88.

\bibitem[{Kosiorek et~al.(2021)Kosiorek, Strathmann, Zoran, Moreno, Schneider,
  Mokr{\'a}, and Rezende}]{kosiorek2021nerf}
Kosiorek, A.~R.; Strathmann, H.; Zoran, D.; Moreno, P.; Schneider, R.;
  Mokr{\'a}, S.; and Rezende, D.~J. 2021.
\newblock Nerf-vae: A geometry aware 3d scene generative model.
\newblock In \emph{International Conference on Machine Learning}, 5742--5752.
  PMLR.

\bibitem[{Lakshminarayanan, Pritzel, and Blundell(2017)}]{deep_ensemble}
Lakshminarayanan, B.; Pritzel, A.; and Blundell, C. 2017.
\newblock Simple and scalable predictive uncertainty estimation using deep
  ensembles.
\newblock \emph{Advances in neural information processing systems}, 30.

\bibitem[{Lee et~al.(2022)Lee, Chen, Wang, Liniger, Kumar, and
  Yu}]{lee2022uncertainty}
Lee, S.; Chen, L.; Wang, J.; Liniger, A.; Kumar, S.; and Yu, F. 2022.
\newblock Uncertainty Guided Policy for Active Robotic 3D Reconstruction Using
  Neural Radiance Fields.
\newblock \emph{IEEE Robotics and Automation Letters}, 7(4): 12070--12077.

\bibitem[{Li et~al.(2022{\natexlab{a}})Li, Li, Zhao, Zhu, and
  Lin}]{Li2022RTNeRFRO}
Li, C.; Li, S.; Zhao, Y.; Zhu, W.; and Lin, Y. 2022{\natexlab{a}}.
\newblock RT-NeRF: Real-Time On-Device Neural Radiance Fields Towards Immersive
  AR/VR Rendering.
\newblock \emph{2022 IEEE/ACM International Conference On Computer Aided Design
  (ICCAD)}, 1--9.

\bibitem[{Li et~al.(2023)Li, Qiao, Chen, Jatavallabhula, Lin, Jiang, and
  Gan}]{Li2023PACNeRFPA}
Li, X.; Qiao, Y.-L.; Chen, P.~Y.; Jatavallabhula, K.~M.; Lin, M.; Jiang, C.;
  and Gan, C. 2023.
\newblock PAC-NeRF: Physics Augmented Continuum Neural Radiance Fields for
  Geometry-Agnostic System Identification.
\newblock \emph{ArXiv}, abs/2303.05512.

\bibitem[{Li et~al.(2021)Li, Li, Sitzmann, Agrawal, and Torralba}]{Li20213DNS}
Li, Y.; Li, S.; Sitzmann, V.; Agrawal, P.; and Torralba, A. 2021.
\newblock 3D Neural Scene Representations for Visuomotor Control.
\newblock \emph{ArXiv}, abs/2107.04004.

\bibitem[{Li et~al.(2022{\natexlab{b}})Li, Li, Ma, Zhang, Chen, and
  Zhu}]{Li2022READLN}
Li, Z.; Li, L.; Ma, Z.; Zhang, P.; Chen, J.; and Zhu, J.-Z. 2022{\natexlab{b}}.
\newblock READ: Large-Scale Neural Scene Rendering for Autonomous Driving.
\newblock \emph{ArXiv}, abs/2205.05509.

\bibitem[{Malinin and Gales(2018)}]{malinin2018predictive}
Malinin, A.; and Gales, M. 2018.
\newblock Predictive uncertainty estimation via prior networks.
\newblock \emph{Advances in neural information processing systems}, 31.

\bibitem[{Martin-Brualla et~al.(2021)Martin-Brualla, Radwan, Sajjadi, Barron,
  Dosovitskiy, and Duckworth}]{nerfw}
Martin-Brualla, R.; Radwan, N.; Sajjadi, M.~S.; Barron, J.~T.; Dosovitskiy, A.;
  and Duckworth, D. 2021.
\newblock Nerf in the wild: Neural radiance fields for unconstrained photo
  collections.
\newblock In \emph{Proceedings of the IEEE/CVF Conference on Computer Vision
  and Pattern Recognition}, 7210--7219.

\bibitem[{Meng et~al.(2021)Meng, Chen, Luo, Wu, Su, Xu, He, and Yu}]{gnerf}
Meng, Q.; Chen, A.; Luo, H.; Wu, M.; Su, H.; Xu, L.; He, X.; and Yu, J. 2021.
\newblock Gnerf: Gan-based neural radiance field without posed camera.
\newblock In \emph{Proceedings of the IEEE/CVF International Conference on
  Computer Vision}, 6351--6361.

\bibitem[{Mildenhall et~al.(2021)Mildenhall, Srinivasan, Tancik, Barron,
  Ramamoorthi, and Ng}]{nerf}
Mildenhall, B.; Srinivasan, P.~P.; Tancik, M.; Barron, J.~T.; Ramamoorthi, R.;
  and Ng, R. 2021.
\newblock Nerf: Representing scenes as neural radiance fields for view
  synthesis.
\newblock \emph{Communications of the ACM}, 65(1): 99--106.

\bibitem[{Miyato et~al.(2018)Miyato, Kataoka, Koyama, and
  Yoshida}]{miyato2018spectral}
Miyato, T.; Kataoka, T.; Koyama, M.; and Yoshida, Y. 2018.
\newblock Spectral normalization for generative adversarial networks.
\newblock \emph{arXiv preprint arXiv:1802.05957}.

\bibitem[{M{\"u}ller et~al.(2022)M{\"u}ller, Evans, Schied, and
  Keller}]{instant-ngp}
M{\"u}ller, T.; Evans, A.; Schied, C.; and Keller, A. 2022.
\newblock Instant neural graphics primitives with a multiresolution hash
  encoding.
\newblock \emph{ACM Transactions on Graphics (ToG)}, 41(4): 1--15.

\bibitem[{Neff et~al.(2021)Neff, Stadlbauer, Parger, Kurz, Mueller, Chaitanya,
  Kaplanyan, and Steinberger}]{neff2021donerf}
Neff, T.; Stadlbauer, P.; Parger, M.; Kurz, A.; Mueller, J.~H.; Chaitanya, C.
  R.~A.; Kaplanyan, A.; and Steinberger, M. 2021.
\newblock DONeRF: Towards Real-Time Rendering of Compact Neural Radiance Fields
  using Depth Oracle Networks.
\newblock In \emph{Computer Graphics Forum}, volume~40, 45--59. Wiley Online
  Library.

\bibitem[{Pan et~al.(2022)Pan, Lai, Song, and Huang}]{active-nerf}
Pan, X.; Lai, Z.; Song, S.; and Huang, G. 2022.
\newblock ActiveNeRF: Learning Where to See with Uncertainty Estimation.
\newblock In \emph{Computer Vision--ECCV 2022: 17th European Conference, Tel
  Aviv, Israel, October 23--27, 2022, Proceedings, Part XXXIII}, 230--246.
  Springer.

\bibitem[{Peng et~al.(2020)Peng, Niemeyer, Mescheder, Pollefeys, and
  Geiger}]{Peng2020ConvolutionalON}
Peng, S.; Niemeyer, M.; Mescheder, L.~M.; Pollefeys, M.; and Geiger, A. 2020.
\newblock Convolutional Occupancy Networks.
\newblock \emph{ArXiv}, abs/2003.04618.

\bibitem[{Poggi et~al.(2020)Poggi, Aleotti, Tosi, and
  Mattoccia}]{poggi2020uncertainty}
Poggi, M.; Aleotti, F.; Tosi, F.; and Mattoccia, S. 2020.
\newblock On the uncertainty of self-supervised monocular depth estimation.
\newblock In \emph{Proceedings of the IEEE/CVF Conference on Computer Vision
  and Pattern Recognition}, 3227--3237.

\bibitem[{Ran et~al.(2023)Ran, Zeng, He, Chen, Li, Chen, Lee, and
  Ye}]{ran2023neurar}
Ran, Y.; Zeng, J.; He, S.; Chen, J.; Li, L.; Chen, Y.; Lee, G.; and Ye, Q.
  2023.
\newblock NeurAR: Neural Uncertainty for Autonomous 3D Reconstruction with
  Implicit Neural Representations.
\newblock \emph{IEEE Robotics and Automation Letters}.

\bibitem[{Ran et~al.(2022)Ran, Zeng, He, Li, Chen, Lee, Chen, and
  Ye}]{ran2022neurar}
Ran, Y.; Zeng, J.; He, S.; Li, L.; Chen, Y.; Lee, G.; Chen, J.; and Ye, Q.
  2022.
\newblock NeurAR: Neural Uncertainty for Autonomous 3D Reconstruction.
\newblock \emph{arXiv preprint arXiv:2207.10985}.

\bibitem[{Schwarz et~al.(2020)Schwarz, Liao, Niemeyer, and
  Geiger}]{schwarz2020graf}
Schwarz, K.; Liao, Y.; Niemeyer, M.; and Geiger, A. 2020.
\newblock Graf: Generative radiance fields for 3d-aware image synthesis.
\newblock \emph{Advances in Neural Information Processing Systems}, 33:
  20154--20166.

\bibitem[{Shen et~al.(2022)Shen, Agudo, Moreno-Noguer, and Ruiz}]{cf-nerf}
Shen, J.; Agudo, A.; Moreno-Noguer, F.; and Ruiz, A. 2022.
\newblock Conditional-flow nerf: Accurate 3d modelling with reliable
  uncertainty quantification.
\newblock In \emph{Computer Vision--ECCV 2022: 17th European Conference, Tel
  Aviv, Israel, October 23--27, 2022, Proceedings, Part III}, 540--557.
  Springer.

\bibitem[{Shen et~al.(2021)Shen, Ruiz, Agudo, and Moreno-Noguer}]{s-nerf}
Shen, J.; Ruiz, A.; Agudo, A.; and Moreno-Noguer, F. 2021.
\newblock Stochastic neural radiance fields: Quantifying uncertainty in
  implicit 3d representations.
\newblock In \emph{2021 International Conference on 3D Vision (3DV)}, 972--981.
  IEEE.

\bibitem[{Straub et~al.(2019)Straub, Whelan, Ma, Chen, Wijmans, Green, Engel,
  Mur-Artal, Ren, Verma, Clarkson, Yan, Budge, Yan, Pan, Yon, Zou, Leon,
  Carter, Briales, Gillingham, Mueggler, Pesqueira, Savva, Batra, Strasdat,
  Nardi, Goesele, Lovegrove, and Newcombe}]{replica}
Straub, J.; Whelan, T.; Ma, L.; Chen, Y.; Wijmans, E.; Green, S.; Engel, J.~J.;
  Mur-Artal, R.; Ren, C.; Verma, S.; Clarkson, A.; Yan, M.; Budge, B.; Yan, Y.;
  Pan, X.; Yon, J.; Zou, Y.; Leon, K.; Carter, N.; Briales, J.; Gillingham, T.;
  Mueggler, E.; Pesqueira, L.; Savva, M.; Batra, D.; Strasdat, H.~M.; Nardi,
  R.~D.; Goesele, M.; Lovegrove, S.; and Newcombe, R. 2019.
\newblock The {R}eplica Dataset: A Digital Replica of Indoor Spaces.
\newblock \emph{arXiv preprint arXiv:1906.05797}.

\bibitem[{Sun et~al.(2022)Sun, Chen, Wang, Li, Averbuch-Elor, Zhou, and
  Snavely}]{Sun2022Neural3R}
Sun, J.; Chen, X.; Wang, Q.; Li, Z.; Averbuch-Elor, H.; Zhou, X.; and Snavely,
  N. 2022.
\newblock Neural 3D Reconstruction in the Wild.
\newblock \emph{ACM SIGGRAPH 2022 Conference Proceedings}.

\bibitem[{Teye, Azizpour, and Smith(2018)}]{teye2018bayesian}
Teye, M.; Azizpour, H.; and Smith, K. 2018.
\newblock Bayesian uncertainty estimation for batch normalized deep networks.
\newblock In \emph{International Conference on Machine Learning}, 4907--4916.
  PMLR.

\bibitem[{Ulyanov, Vedaldi, and Lempitsky(2016)}]{ulyanov2016instance}
Ulyanov, D.; Vedaldi, A.; and Lempitsky, V. 2016.
\newblock Instance normalization: The missing ingredient for fast stylization.
\newblock \emph{arXiv preprint arXiv:1607.08022}.

\bibitem[{Vora et~al.(2021)Vora, Radwan, Greff, Meyer, Genova, Sajjadi, Pot,
  Tagliasacchi, and Duckworth}]{Vora2021NeSFNS}
Vora, S.; Radwan, N.; Greff, K.; Meyer, H.; Genova, K.; Sajjadi, M. S.~M.; Pot,
  E.; Tagliasacchi, A.; and Duckworth, D. 2021.
\newblock NeSF: Neural Semantic Fields for Generalizable Semantic Segmentation
  of 3D Scenes.

\bibitem[{Wang et~al.(2021)Wang, Liu, Liu, Theobalt, Komura, and
  Wang}]{wang2021neus}
Wang, P.; Liu, L.; Liu, Y.; Theobalt, C.; Komura, T.; and Wang, W. 2021.
\newblock Neus: Learning neural implicit surfaces by volume rendering for
  multi-view reconstruction.
\newblock \emph{arXiv preprint arXiv:2106.10689}.

\bibitem[{Winkler et~al.(2019)Winkler, Worrall, Hoogeboom, and Welling}]{cnf}
Winkler, C.; Worrall, D.; Hoogeboom, E.; and Welling, M. 2019.
\newblock Learning likelihoods with conditional normalizing flows.
\newblock \emph{arXiv preprint arXiv:1912.00042}.

\bibitem[{Yang et~al.(2019)Yang, Huang, Hao, Liu, Belongie, and
  Hariharan}]{yang2019pointflow}
Yang, G.; Huang, X.; Hao, Z.; Liu, M.-Y.; Belongie, S.; and Hariharan, B. 2019.
\newblock Pointflow: 3d point cloud generation with continuous normalizing
  flows.
\newblock In \emph{Proceedings of the IEEE/CVF international conference on
  computer vision}, 4541--4550.

\bibitem[{Yariv et~al.(2021)Yariv, Gu, Kasten, and Lipman}]{yariv2021volume}
Yariv, L.; Gu, J.; Kasten, Y.; and Lipman, Y. 2021.
\newblock Volume rendering of neural implicit surfaces.
\newblock \emph{Advances in Neural Information Processing Systems}, 34:
  4805--4815.

\bibitem[{Yu et~al.(2022)Yu, Peng, Niemeyer, Sattler, and
  Geiger}]{Yu2022MonoSDFEM}
Yu, Z.; Peng, S.; Niemeyer, M.; Sattler, T.; and Geiger, A. 2022.
\newblock MonoSDF: Exploring Monocular Geometric Cues for Neural Implicit
  Surface Reconstruction.
\newblock \emph{ArXiv}, abs/2206.00665.

\bibitem[{Zhang et~al.(2020)Zhang, Riegler, Snavely, and
  Koltun}]{zhang2020nerf++}
Zhang, K.; Riegler, G.; Snavely, N.; and Koltun, V. 2020.
\newblock Nerf++: Analyzing and improving neural radiance fields.
\newblock \emph{arXiv preprint arXiv:2010.07492}.

\bibitem[{Zhi et~al.(2021)Zhi, Laidlow, Leutenegger, and
  Davison}]{zhi2021place}
Zhi, S.; Laidlow, T.; Leutenegger, S.; and Davison, A.~J. 2021.
\newblock In-place scene labelling and understanding with implicit scene
  representation.
\newblock In \emph{Proceedings of the IEEE/CVF International Conference on
  Computer Vision}, 15838--15847.

\end{thebibliography}

\end{document}